\documentclass{article}
\pdfoutput=1

\PassOptionsToPackage{numbers}{natbib}
\usepackage[final]{nips_2017}

\usepackage[utf8]{inputenc} % allow utf-8 input
\usepackage[T1]{fontenc}    % use 8-bit T1 fonts
\usepackage{hyperref}       % hyperlinks
\usepackage{url}            % simple URL typesetting
\usepackage{booktabs}       % professional-quality tables
\usepackage{amsfonts}       % blackboard math symbols
\usepackage{nicefrac}       % compact symbols for 1/2, etc.
\usepackage{microtype}      % microtypography
\usepackage[pdftex]{graphicx}
\DeclareGraphicsExtensions{.pdf}
\usepackage[font=footnotesize]{caption}
\usepackage{xcolor}
\usepackage{mathtools}
\usepackage{bm}
\usepackage{wrapfig}

\usepackage{subcaption}
\usepackage{cleveref}
\newcommand{\labelphantom}[1]{\parbox{0pt}{\phantomsubcaption\label{#1}}}

\title{
  CortexNet: a Generic Network Family for \\
  Robust Visual Temporal Representations
}

\author{
  Alfredo Canziani \& Eugenio Culurciello\\
  Weldon School of Biomedical Engineering\\
  Purdue University\\
  \texttt{\{canziani,euge\}@purdue.edu}
}

\begin{document}

\maketitle

\begin{abstract}
  In the past five years we have observed the rise of incredibly well performing feed-forward neural networks trained supervisedly for vision related tasks. %In my opinion remove this sentence
  These models have achieved super-human performance on object recognition, localisation, and detection in still images.
  However, there is a need to identify the best strategy to employ these networks with temporal visual inputs and obtain a robust and stable representation of video data.
  Inspired by the human visual system, we propose a deep neural network family, CortexNet, which features not only bottom-up feed-forward connections, but also it models the abundant top-down feedback and lateral connections, which are present in our visual cortex.
  We introduce two training schemes --- the unsupervised MatchNet and weakly supervised TempoNet modes --- where a network learns how to correctly anticipate a subsequent frame in a video clip or the identity of its predominant subject, by learning egomotion clues and how to automatically track several objects in the current scene.
  Find the project website at \texttt{tinyurl.com/CortexNet}.
\end{abstract}

\section{Introduction}

We have recently seen a wide and steady release of state-of-the-art feed-forward deep convolutional neural networks for vision related tasks \citep{canziani2016analysis}. %You don't need active voice here. You are telling facts
These models have reached, and then also surpassed the human-level performance of object recognition \citep{he2015delving} in the ImageNet classification challenge \citep{russakovsky2015imagenet}.
Currently, these models are trained end-to-end, using strong supervision.
This means that large collections of annotated still-images are fed to the networks, the gradient of a cross entropy loss function with respect to the model parameters is computed with back-propagation, and gradient descent is used to minimise the error between the prediction and the ground truth. %You need active voice here though. You are telling a process
We then want to utilise these models for real life applications, feeding them with a stream of video frames, and expecting them to behave similarly well on live data, but this is not often the case.

Furthermore, these models are highly susceptible to inputs corrupted by adversarial noise \citep{nguyen2015deep}.
Such inputs are made up of small carefully designed perturbations, which are invisible to the normal human vision.
For some extent, we can attribute the temporal prediction instability of the feed-forward models to the natural occurrence of adversarial noise.
Arguably, our visual system is immune to such temporal perturbations, because in the early years of an individual it has been ``trained to see'' by performing tracking on specific objects \citep{aoa-infant} with sporadic parental weak supervision, and not from a large collection of static annotated flash cards.

Therefore, we propose CortexNet, a neural network family which not only models the bottom-up feed-forward connections in our visual system, but also employs delayed modulatory feedback with lateral connections, in order to learn end-to-end a more robust representation of natural temporal visual inputs.
We train our models either unsupervisedly or with weak sparse annotations, through leveraging of the temporal coherence which is present among the frames of a natural video clip.
Our models show short-term reliable next frames prediction by (1) compensating for the camera egomotion, (2) learning the trajectory of the object present in the current scene, and (3) focussing on one object at the time.
Our preliminary results indicate that the network develops an internal salient and attention mechanism.
This leads to an effective internal representation of our reality, which demonstrates a superior and more robust network class.

\section{Related work}

Two main types of work are related to our research direction.
The first type uses the natural temporal order of frames in a video as signal --- or self-supervised pretext --- to train neural networks and learn static visual representations.
This eliminates the need of expensive large annotated data sets.
On the other hand, the second aims to learn a temporal visual representation directly from the video data itself, by means of future and past frame reconstruction.

\subsection{Learning static visual representations from videos}

Exploiting the motion present in video data to learn visual representations is a prevalent approach used by self-supervised and unsupervised learning techniques since the frames' temporal coherency comes to us without any cost. %gratis is a really wrong and obscure word here
\citet{wang2015unsupervised} use a triplet loss to train a network so that it learns to differentiate patches belonging to a given tracked object against the patches that do not.
In this case, tracking is performed with non trivial algorithms, and it is used to generate positive and negative training samples.
In our work, we delegate to the network itself for computing any necessary operation directly on the source video data, and thus training the whole architecture end-to-end.
Similarly, the model of \citet{vondrick2016anticipating} demonstrates prediction of the future embedding for a video sequence, given only the current frame, and without a system state.
Nevertheless, predicting future representations is ill-posed \emph{per se}, given that the only real ground truth is the unprocessed reality that is available to the model.
\citet{pathak2016learning} exploit motion generated segmentation maps for training a neural network to segment objects from a single frame.
Once more, even though videos are utilised, the network is still operating in a feed-forward-only configuration and does not exploit temporal cues.

\subsection{Learning dynamic visual representations}

Remarkably, the most relevant previous work is the seminal NIPS `96 paper by \citet{softky1996unsupervised} which uses a three-layer spiking feed-forward and feedback network --- using kernels of 4 units, stride of 2, max pooling, and multiplicative signals combiner --- in order to predict the next frame in a natural video. %Surprisingly is the incorrect fit here
Our model can be seen as a conversion and upgrade of \citeauthor{softky1996unsupervised}'s in a deep learning key, where the main building blocks are strided (de)convolutions, non-linearities, and additive signal mergers.
\citet{srivastava2015unsupervised} propose to learn to reconstruct the future and past sequences of frames or their representations, by utilising an encoder-decoder recurrent network scheme, fed with 1D unrolled images or embeddings of a feed-forward convolutional net.
We propose a model that is aware of temporal variations of its input pixels, and is able to perceive motion in its early layers, operating directly on spatial inputs.
The spatio-temporal video auto-encoder of \citet{patraucean2015spatio} is able to predict the next frame in a clip, by using a combination of spatio-recurrent, optical flow, smoothness penalty, grid generator, and sampler modules.
Instead of drawing inspiration from standard video encoders and compression schemes, we are motivated by biologically plausible simpler alternatives.

Finally --- inspired by the neuroscientific predictive coding theory introduced by \citet{rao1997dynamic} and expanded by \citet{friston2008hierarchical} --- \citet{chalasani2013deep} and \citet{lotter2016deep} propose their respective stacks of hand-crafted modules.
Instead, we do not choose to engineer our modules, but to learn the necessary operations directly from the input data.
We believe that a more generic and simpler architecture structure will provide the ground for an easier comparison and analysis of the learnt internal representation.
In addition, we have experimented widely with \citeauthor{lotter2016deep}'s PredNet with multiple supervised tasks, however, we were unable to identify an ultimate strategy to utilise the learnt distributed representation.
Therefore, we introduce a new network family which could be trained with several losses (although we present here just three of them), which allow us to obtain a usable model, with direct practical impact on working applications.

\section{Model architecture family}

We introduce a new family of networks which models not only feed-forward (bottom-up) but also lateral (horizontal) and feedback (top-down) connections between cortical areas of the visual system, that have been shown to provide perceptual context modulation (attention) \citep{lamme1998feedforward}.
As depicted in \cref{fig:model_02_and_blow-ups:model}, the model architecture is composed by $L$ discriminative and generative blocks $D_{1:L}$ and $G_{1:L}$.
Each pair $(D_n, G_n)$ is expected to model a specific cortical area of the human visual system.
$G_n$, fed with the superposition of top-down and bottom-up (also called \emph{residual}) projections, provides a modulatory input to its correspondent $D_n$, based on the previous time step (\emph{i.e.}\ the $[t-1]$ connections in \cref{fig:model_02_and_blow-ups:model}).
% At the moment of writing, there is concurrent effort to map each modelled cortical area to fMRI scans \citep{HG}.

A blow-up of $D_n$ and $G_n$ blocks are shown respectively in \cref{fig:model_02_and_blow-ups:D,fig:model_02_and_blow-ups:G}.
We can observe that the branching and superposition operations happen right after the spatial projection onto the (de)convolutional kernels, as it has been proven to be more promising \citep{he2016identity}.
The (de)convolutional kernels are all $3 \times 3$ with a stride of $2$ and four-sided padding of $1$, and the number of maps are $3, 32, 64, 128, 256$, and additional layers have all $256$ features.
For $D_{2:L}$ blocks, the number of input maps is considered to be doubled due to the concatenation module.
Finally, the feedback connections are initialised at time $t = 0$, to appropriately sized zero tensors.

\begin{figure}[!t]
  \centering
  \labelphantom{fig:model_02_and_blow-ups:model}
  \labelphantom{fig:model_02_and_blow-ups:D}
  \labelphantom{fig:model_02_and_blow-ups:G}
  \labelphantom{fig:model_02_and_blow-ups:embedding}
  \includegraphics{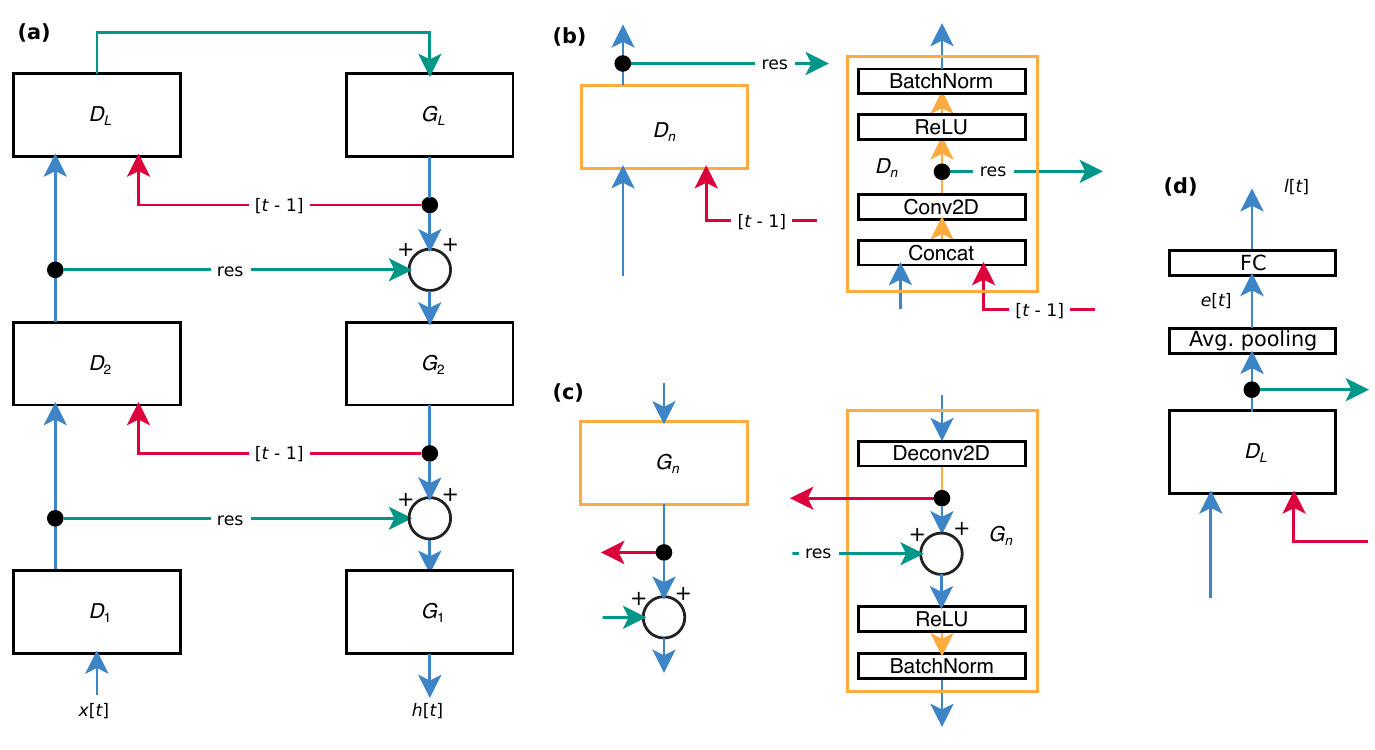}
  \caption{
    \textbf{(a) Model architecture, (b) discriminative and (c) generative blocks blow-ups, and (d) model's embedding and logits.}
    (a) The model architecture features two types of modules, called discriminative $D$ and generative $G$, which are linked together through feed-forward, lateral, and feedback connections.
    Vertical connections are drawn in blue, temporal feedback connections in red and residual lateral connections in green.
    More details about the $D$ and $G$ blocks are shown in the two blow-ups (b, c), where we can notice that the branching and superposition operations take place right after the (de)convolutional modules.
    $D_1$ and $G_1$ blocks do not have the concatenation and the superposition modules respectively, and they can be thought as the sensory input and motor output interfaces.
  (d) Definition of model's embedding $e[t]$ and logits $l[t]$.
  }
  \label{fig:model_02_and_blow-ups}
\end{figure}

\section{Training frameworks}

We apply two schemes to observe two different learning paradigms: MatchNet (unsupervised training configuration), and TempoNet (weakly supervised training configuration) .
Both schemes feed the network through the $D_1$ sensory input interface with batches (of size $\beta = 20$) of video sequences (with minor side scaled to $256$ pixels, and square centre cropped) $x[t]$, in multiple temporal chunks (of length $T = 10$ frames).
We consider all our videos as they were linearly concatenated into a long clip of $N$ frames, and then reshaped it into a rectangle of height $\beta$ and width $\lceil N / \beta \rceil$.
The potentially remaining empty positions are filled with up to $\beta - 1$ frames from the first video. %What do you mean by the word potentially here?

It was our intention to initially pre-train our model unsupervisedly in MatchNet configuration --- so that it learns the dynamics of the videos present in our data set --- followed by using it as a TempoNet with minimal effort and supervision.
As we will see from both the \cref{sec:results:unsup,sec:results:sup}, these two schemes seem to be mutually exclusive when we operate on the pixel-space.
Similar findings are reported by \citet{neverova2017predicting} which are in contrast with the findings of \citet{lotter2016deep}.

Here we define four loss functions, which allow us to explore the model affinity to learn video features.
The future-\underline{m}atching $\mathcal{L}_\mu$ and \underline{r}eplica-checking $\mathcal{L}_\rho$ loss functions are both defined as:
% equation
\begin{equation}\label{eq:MSE}
  \mathcal{L}_\mu(a, b) =\mathcal{L}_\rho(a, b) = \mathrm{MSE}(a, b) \equiv \frac{1}{\#a} \sum (a - b)^2
\end{equation}
% equation
where $a$ and $b$ are two same sized tensors (or tensorial batches), MSE stands for mean squared error, $\#$ is the elements count operator, and the summation is performed across every dimension.
The \underline{t}emporal-stabilisation $\mathcal{L}_\tau$ and \underline{p}eriodic-classification $\mathcal{L}_\pi$ loss functions are defined as:
% equation
\begin{equation}\label{eq:CE}
  \mathcal{L}_\tau(l, c, \bm{w}) =\mathcal{L}_\pi(l, c, \bm{w}) = \mathrm{CE}(l, c, \bm{w}) \equiv - w_c \log[\mathrm{softmax}(l)]\bigr\rvert_c
\end{equation}
% equation
where $l$ represents our logits (spatial average pooling of $D_L$'s output, named \emph{embedding}, which underwent a final linear transformation, in order to have $K$ output dimensions) shown in \cref{fig:model_02_and_blow-ups:embedding}, $c$ is the correct class label index associated to the current video frame, $\bm{w} \in \mathbb{R}^K$ are the class balancing weights, CE stands for cross entropy, and $\mathrm{softmax}(l)[k] \equiv \exp(l[k])\big/\sum_{j=1}^K\exp(l[j])$.
We average along the batch dimension, if batches are used.
The system's training loss $\mathcal{L}$ is defined as a linear combination of the \underline{m}atching $\mathcal{L}_\mu$, \underline{t}emporal $\mathcal{L}_\tau$, and \underline{p}eriodic $\mathcal{L}_\pi$ and, more precisely, as:
% equation
\begin{equation}\label{eq:system_loss}
  \mathcal{L} = \mu \mathcal{L}_\mu + \tau \mathcal{L}_\tau + \pi \mathcal{L}_\pi
\end{equation}
% equation
while we use the \underline{r}eplica $\mathcal{L}_\rho$ to monitor the training health.
We use the Greek letters as mnemonics for the respective losses, \emph{i.e.}\ $\mu$-\underline{m}atching, $\rho$-\underline{r}eplica, $\tau$-\underline{t}emporal, and $\pi$-\underline{p}eriodic loss.

\begin{figure}[!t]
  \centering
  \labelphantom{fig:training_mode:MatchNet}
  \labelphantom{fig:training_mode:TempoNet}
  \includegraphics{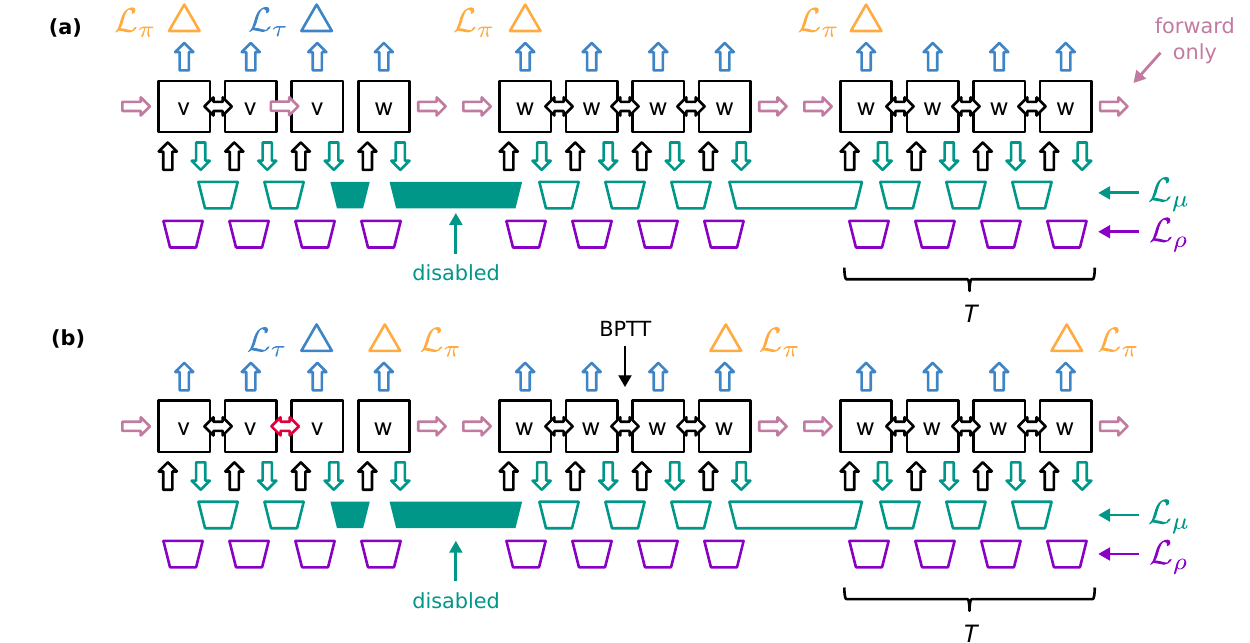}
  \caption{
    \textbf{(a) MatchNet and (b) TempoNet training configurations.}
    Each black square box represents an instance of the model, which is replicated over time with parameter sharing.
    The green contour trapezoids represent the computation of the matching loss $\mathcal{L}_\mu$.
    The purple trapezoids compute the replica matching loss $\mathcal{L}_\rho$, which is used for monitoring the training health.
    The couples of filled green trapezoids are disabled because (1) we reached the last frame of video $v$ --- and therefore, the model cannot match any new frame for the same video --- and (2) the second frame prediction for video $w$ would be erroneous --- since the state has just been reset. %The couples of filled green trapezoids, I don't understand this?
    The double headed arrows indicate where back-propagation-though-time is performed, while the right-pointing pink arrows show when the state is propagated forward.
    However, there is no gradient propagated in the opposite direction (this happens when we reach the end of video $v$, or we start a new BPTT temporal chunk).
    The blue and yellow triangles represent the computation of the classification losses $\mathcal{L}_\tau$ and $\mathcal{L}_\pi$ respectively, fed with a linear transformation of the model embedding $e[t]$ (spatial average pooling of $D_L$'s output, see \cref{fig:model_02_and_blow-ups:embedding}).
  }
  \label{fig:training_mode}
\end{figure}

\subsection{MatchNet mode} \label{sec:MatchNet_mode}

In MatchNet mode, we unsupervisedly train the model in order to minimise $\mathcal{L}_\mu(h[t], x[t+1])$, \emph{i.e.}\ matching the next frame appearance within the same video clip.
Prediction of $x_v[2]$ (second frame of a generic video $v$) and $x_v[T_v + 1]$ or $x_w[1]$ (one frame after the last one for video $v$ or the first frame of video $w$) are disabled (see \cref{fig:training_mode:MatchNet}), since they would be erroneous due to state reset or missing data.
Our expectation is the ability to predict the future scene, \emph{i.e.}\ $h[t] = \hat x[t+1]$, that would build an unsupervised internal representation of the video dynamics, which we could later exploit for other tasks.
We use then $\mathcal{L}_\rho(h[t], x[t])$ --- representing the level of similarity of $G_1$'s output and the current input frame --- in order to monitor the health of the training procedure, and check whether we are effectively matching the next frame or replicating the current one.

We also employ the $\mathcal{L}_\tau(e[\bar T_v], i[v], \bm{1})$ every time a video $v$ reaches its last frame $x_v[T_v]$ (see blue triangle in \cref{fig:training_mode:MatchNet}), where $e[\bar T_v]$ is the last model embedding for video $v$ (see note\footnote{
% footnote
  Here we have used an implicit indexing conversion, from the number of frames $T_v$ of video $v$ and the location $\bar T_v$ where $v$ reaches its last frame in the batched input data $x[t]$.
% footnote
}), $i[v]$ indicates $v$'s index (\emph{e.g.}\ if our data set has $m$ videos, then $1 \leq i \leq m$), $\bm{1}$ is the one-vector, and $x_v[t]$ represents the video $v$ which has $T_v$ frames.
We use $\mathcal{L}_\pi(e[nT + 1], i[v], \bm{1}), n \in \mathbb{N}$ only to monitor the prediction loss ($\pi = 0$ in \cref{eq:system_loss}), periodically, at every temporal chunk for all the videos in the current batch (see yellow triangles in \cref{fig:training_mode:MatchNet}).
While we use $\mathcal{L}_\mu$ to train the model via back-propagation-though-time (BPTT) and learn the video dynamics, we utilise $\mathcal{L}_\tau$ in a static manner, which means that the gradient is not sent backwards to early time steps (notice the pink right-pointing arrow, instead of a double headed one, in correspondence of $x_v[T_v]$ in \cref{fig:training_mode:MatchNet}).

In order to validate our model performance, we split every video $x_v[t]$ into $x_v^\mathrm{train}[t] = x_v[1 : T_v - 60]$ and $x_v^\mathrm{val}[t] = x_v[T_v - 59 : T_v]$.
Given that the clips' frame rate is $30\,\mathrm{Hz}$, this means that we use the last two seconds of each video as validation data.

\subsection{TempoNet mode}

In TempoNet mode we train our model with weak supervision over object class recognition through the periodic loss $\mathcal{L}_\pi(e[nT], c[v], \bm{w}), n \in \mathbb{N}^+$, where $c[v]$ represents the object class of each video $v$ in the current batch.
We compute each $\bm{w}$'s component $w[k]$ as:
% equation
\begin{equation}\label{eq:CE_balancing}
  w[k] = \frac{1}{K} \sum_{j=1}^K m[j] \Bigg/ m[k]
\end{equation}
% equation
where $\bm{m} \in \mathbb{R}^K$ represents the number of samples for each of the $K$ classes.
Moreover, we used $\mathcal{L}_\tau(e[nT], c[v], \bm{w}), n \in \mathbb{N}^+$ to monitor only the training status ($\tau = 0$ in \cref{eq:system_loss}).

To avoid the situation where all the $e[nT]$ across the batch are held to a constant value, we decided to implement the following data feeding strategy.
Each video $x_v[t]$ is split into $S = 5$ subsampled videos $x_v^s[t] = x_v[s + nS], 1 \leq s \leq S, n = t - 1 \in \mathbb{N}$; the training set is made of $x_v^{1:S-1}[t]$ splits, and the validation gets the remaining $x_v^S[t]$.
In this way the sampling rate goes from $30\,\mathrm{Hz}$ to $6\,\mathrm{Hz}$, and the average training video length goes from $300$ ($\simeq 354-60$) of MatchNet mode to $70$ ($\simeq 354 / 5$).
This means given that $T = 10$, we are going to observe an average of $2.9$ ($\simeq \beta T / 70$) video changes per temporal chunk \emph{vs.}\ $0.7$ ($\simeq \beta T / 300$) of the previous training scheme, which would have caused the network to converge to an interesting unstable equilibrium point, given that our model is a dynamic non-linear system.

\section{Experiments and results}

In this section, we present the major results and relative experimental settings that showcase the performance of the CortexNet architecture family.
For our experiments we have used the e-Lab Video Data Set (e-VDS35) \citep{e-VDS}, a growing collection of currently 970 clips of roughly 10 seconds ($\simeq 354$ frames) each, capturing one of 35 common life main objects from different perspectives.
Duration outliers have been removed by extracting the two-sided $95\%$ confidence interval of a Student's $t$-distribution fitted on the video lengths population.
This means removing clips shorter than $144$ frames and trimming the ones longer than $564$ frames.
The project source code can be found at \citep{CortexNet}.

\pagebreak
\subsection{Unsupervised learning} \label{sec:results:unsup}

%\begin{figure}[!t]
\begin{wrapfigure}{r}{0.5\textwidth}\vspace{-20pt}
  \centering
  \labelphantom{fig:MatchNet:images}
  \labelphantom{fig:MatchNet:chart}
  \includegraphics[width=.5\textwidth]{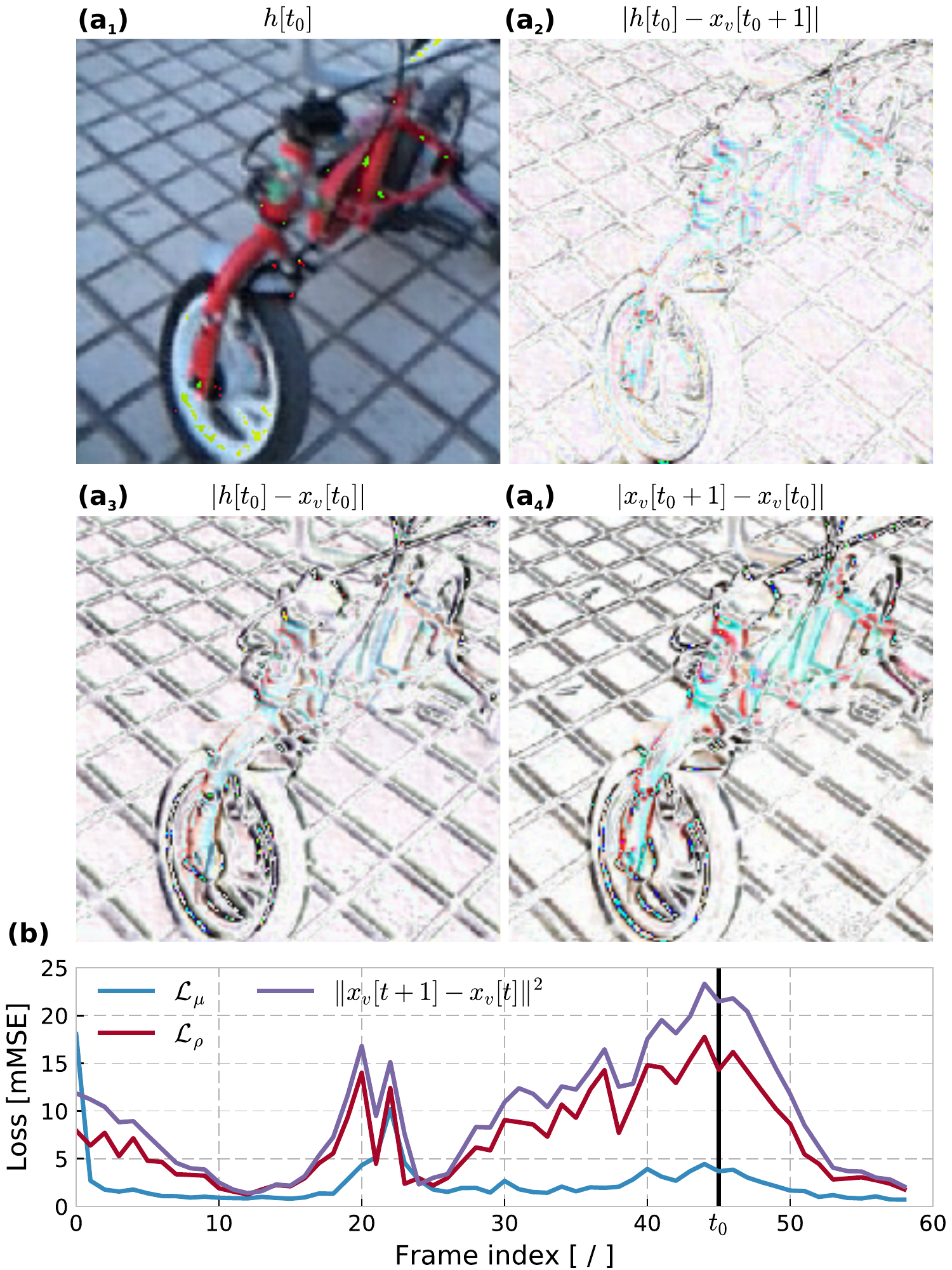}
  \caption{
    \textbf{(a) $\bm{h[t_0]}$ and distances between $\bm{h[t_0]}$, $\bm{x_v[t_0]}$ and $\bm{x_v[t_0+1]}$, (b) MSE losses \emph{vs.}\ time.}\\
    Initially, the network's output $h[t=0]$ is much closer to the current input ($\mathcal{L}_\mu > \mathcal{L}_\rho$).
    As soon as two frames have been fed, the network locks on the temporal visual clues, and predicts the next frame reliably  ($\mathcal{L}_\mu < \mathcal{L}_\rho$) until the camera auto-focus kicks in at $t=20$, where the input gets completely blurred.
    After that, the network keeps predicting its egomotion, even at the highest panning speed at $t = t_0$.
    In the upper part, we can see how the network's output $h[t]$, current and next frames compare to each other at time index $t_0$.
  Animation available at \citep{CortexNet}.
  }
  \label{fig:MatchNet}\vspace{-20pt}
%\end{figure}
\end{wrapfigure}

In MatchNet mode, we train our model solely on unlabelled data and exploit the inherent data statistics as learning signals.
More precisely, as explained in \cref{sec:MatchNet_mode}, we feed the network batches of consecutive frames and utilise $\mathcal{L}_\mu$ to predict the next frame and $\mathcal{L}_\tau$ to identify which video we have just finished processing, with no reference to the object class it belongs to.
In this way, we leverage only the intrinsic characteristics of our data and supervision is reduced to its minimum.
We trained a four-layer CortexNet, with a 970-dimensional output logits on top of $D_4$.
We used a momentum of $0.9$, a weight decay of $10^{-4}$, and an initial learning rate of $0.1$, and had it decay by a factor of $10$ every 10 epochs, for a total of 30 epochs.
We set $\mu = 1$, $\tau = 0.01$ and $\pi = 0$ in \cref{eq:system_loss}.
We obtained a $\mathcal{L}_\mu^\mathrm{val} = 2.1\,\mathrm{mMSE}$ compared to $\mathcal{L}_\rho^\mathrm{val} = 4.8\,\mathrm{mMSE}$, which means we are more than twice as better to predict the next frame \emph{vs.}\ barely copying the input one.
In \cref{fig:MatchNet} we can see how the model behaves on an interesting video from the validation set.
The purple line in \cref{fig:MatchNet:chart} --- defined as $\mathrm{MSE}(x_v[t+1], x_v[t])$ --- represents the panning speed: the higher its value, and the larger amount of motion has been recorded.
The video clip starts with an initial non-zero panning, it slows down in 12 frames, the camera auto-focusses around frame 20, panning increases to its maximum around frame 44 and then it slows down again.
The model is constantly tracking the different moving elements, predicting successfully the future frame when the input is not corrupted (blurred by the auto focus).

Surprisingly, as soon as the network learns to accurately predict the next frame in our training clips, the output logits becomes constant, and the $\mathcal{L}_\pi = \mathcal{L}_\tau = \log(970) \simeq 6.88\,\mathrm{nit}$.
This indicates that the task of generating the future input appearance is effectively executed by a few of the lower $(D_n, G_n)$ pairs.
Therefore, we investigate whether we can obtain a more useful high level representation with an alternative training strategy.

\subsection{Supervised learning} \label{sec:results:sup}

%Abrupt starting, continue the flow from last section
In TempoNet mode, we exclusively train our model from the top representation through $\mathcal{L}_\pi$, over the object classes with BPTT (note that MatchNet used video indices and static back-prop for classification). %Please reform this first sentence
Although we experimented with several values for $\mu$ in \cref{eq:system_loss}, it had showed no overall effect on the classification task, and we got as good as $\mathcal{L}_\mu^\mathrm{val} = 14.2\,\mathrm{mMSE}$ compared to $\mathcal{L}_\rho^\mathrm{val} = 7.4\,\mathrm{mMSE}$ for $\mu = 10^{-3}$, probably due to the amount of motion introduced by the subsampling.

\begin{figure}[!ht]
  \centering
  \labelphantom{fig:TempoNet:discriminator}
  \labelphantom{fig:TempoNet:CortexNet}
  \labelphantom{fig:TempoNet:saliency}
  \includegraphics[width=1.0\textwidth]{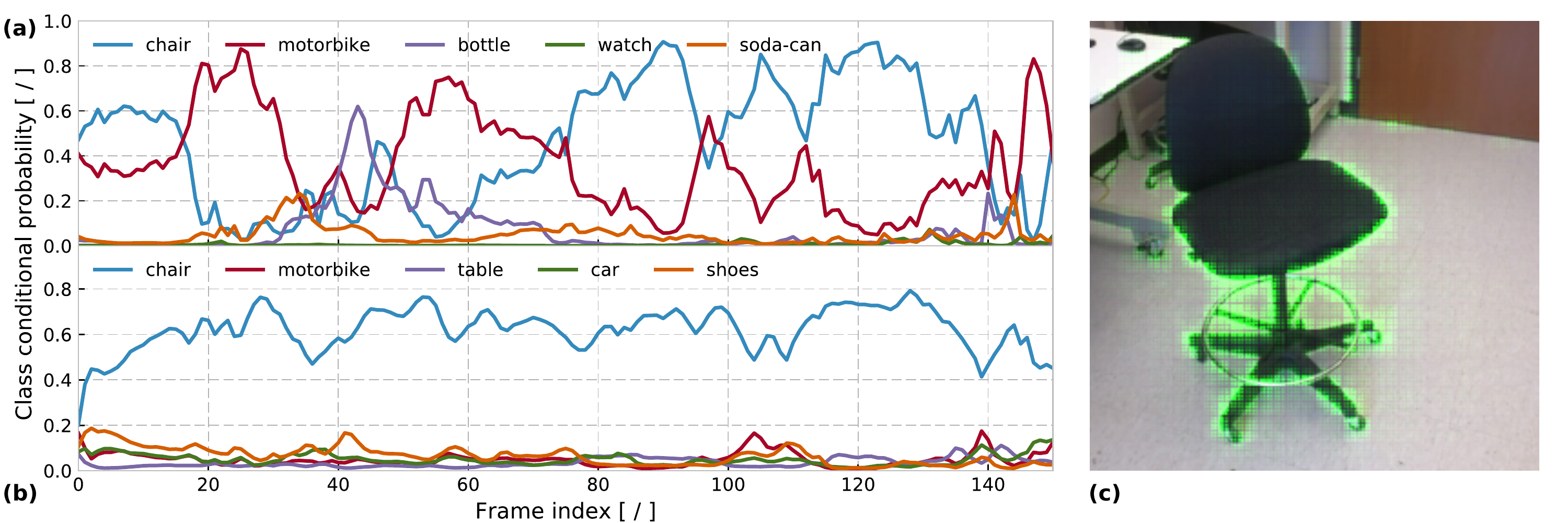}
  \caption{
    \textbf{Probability \emph{vs.}\ time index for (a) discriminator only and (b) full CortexNet architecture, and (c) salient regions highlight.}
    In these charts, spanning 5 seconds, we can see how flickered is the output of a feed-forward architecture (CortexNet discriminative branch only) compared to the corresponding full CortexNet model.
    Even with a temporal varying input, our TempoNet is able to track moving object in the scene, focussing its dynamic attention to it, and steadily predict the correct object identity.
    In order to better understand how the model is able to flawlessly perform such task, we utilised a salient region highlighter algorithm \citep{bojarski2017explaining} to visualise over time, where the network is looking at, and which is included in the snapshot. %Flawlessly seems a wrong fit and overselling to me
  Animation available at \citep{CortexNet}.
  }
  \label{fig:TempoNet}
\end{figure}

We pre-trained the discriminative branch of a six-layer CortexNet with a 33-dimensional output logits on top of $D_6$, on an image data set of 300k images of objects similar to the ones in e-VDS35 \citep{e-VDS}.
This data set is a subset of the Open Images one \citep{krasin2016openimages}.
For the pre-training, we used a momentum of $0.9$, a weight decay of $10^{-4}$, and an initial learning rate of $0.1$, and reduced it by $10$ every $30$ epochs, and trained the feed-forward branch for $90$ epochs.
The performance of the discriminator branch on a video clip is shown in \cref{fig:TempoNet:discriminator}.

Then we added the generative branch, swapped the classifier with a 35-dimensional one, and fine-tuned the whole model on e-VDS35, with $\pi = 10^{-2}$, $\tau = 0$ and $\mu = 0$ (in \cref{eq:system_loss}).
We used a momentum of $0.9$, a weight decay of $10^{-4}$, and an initial learning rate of $0.1$, and reduced it by $\sqrt{10}$ every $10$ epochs, while training the model for a total of $30$ epochs.
The model that we have obtained is now much more robust to temporal perturbations of the input video stream (see \cref{fig:TempoNet:CortexNet}), by adopting a selective attention mechanism to focus and track over time the main object present in the scene.
Additionally, we are using salient-object-finding algorithms \citep{bojarski2017explaining,canziani2015visual} because we want to visualise dynamically the locations where the network is currently paying attention (see snapshot in \cref{fig:TempoNet:saliency}).

\section{Conclusions}
%Please rewrite this section. It's "conclusions", not summary.
In this paper, we introduce a new kind of neural network family, called CortexNet, which not only model the bottom-up feed-forward connections in the human visual system but employs delayed modulatory feedback and lateral connections, in order to learn end-to-end a more robust representation of natural temporal visual inputs. %You have repeated this sentence, please change it up. Tell your objective or gap and then tell what you have done.
We explore an unsupervised and a weakly supervised training strategy to train two models on a custom, object-centric video data set.
We report performance in terms of prediction mean square error and compare it to the input-matching trivial task, and we show also how the new architecture provide a much more stable prediction output on a testing video clip.
Lastly, we observe that the task of predicting a future frame, directly in pixel space, is not complementary to the one of predicting low-frequency labels, such as video index prediction, action recognition or anything that spans several tens of frames in time.

\subsubsection*{Acknowledgements}

This project leveraged the power, speed, and quick implementation time of \emph{PyTorch} for all computationally expensive operations.
It resorted to the illustrating capabilities of the \emph{Inkscape} vector graphics software.
It also explored and visualised data though the \texttt{matplotlib} library combined with the \emph{Jupyter Notebook} interactive computational environment.
This work was partly sponsored by the Office of Naval Research grants N00014-15-1-2791 and N00014-17-1-2225.
We also thank NVIDIA for the donations of graphical processors.

{
  \footnotesize
  \bibliographystyle{myplainnat}
  \bibliography{ref}
}

\end{document}